%% file: iclr2021_conference.tex
\documentclass{article} 
\usepackage{iclr2021_conference,times}

\usepackage{wrapfig}
\usepackage{graphicx}
\usepackage{lipsum}
\usepackage{subfig}
\usepackage[ruled]{algorithm2e}
\input{math_commands.tex}

\usepackage{hyperref}
\usepackage{url}

\title{
Transfer Learning Between Different \\ Architectures Via Weights Injection
}

\author{Maciej A.~Czyzewski \\
Poznan University of Technology, Poznan, Poland\\
\texttt{maciejanthonyczyzewski@gmail.com} \\
}

\iclrfinalcopy
\begin{document}

\maketitle
\lhead{DRAFT} 


\begin{abstract}
This work presents a naive algorithm for parameter transfer
\underline{between} \underline{different} \underline{architectures} with a computationally cheap injection technique (which does not require data).
The primary objective is to speed up the training of neural networks from scratch.
It was found in this study that transferring knowledge from any architecture was superior to Kaiming and Xavier for initialization.
In conclusion, the method presented is found to converge faster, which makes it a drop-in replacement for classical methods.
The method involves:
1) matching: the layers of the pre-trained model with the targeted model;
2) injection: the tensor is transformed into a desired shape.
This work provides a comparison of similarity between the current SOTA
architectures (ImageNet), by utilising TLI (Transfer Learning by Injection)
score.
\end{abstract}

\newcommand{\ComboInjection}{\ensuremath{\text{ComboInjection}}}


\section{Introduction}

\vspace{-2mm}

\begin{wrapfigure}{R}{.5\columnwidth}
    \vspace{-11mm}
    \center\includegraphics[width=\linewidth]{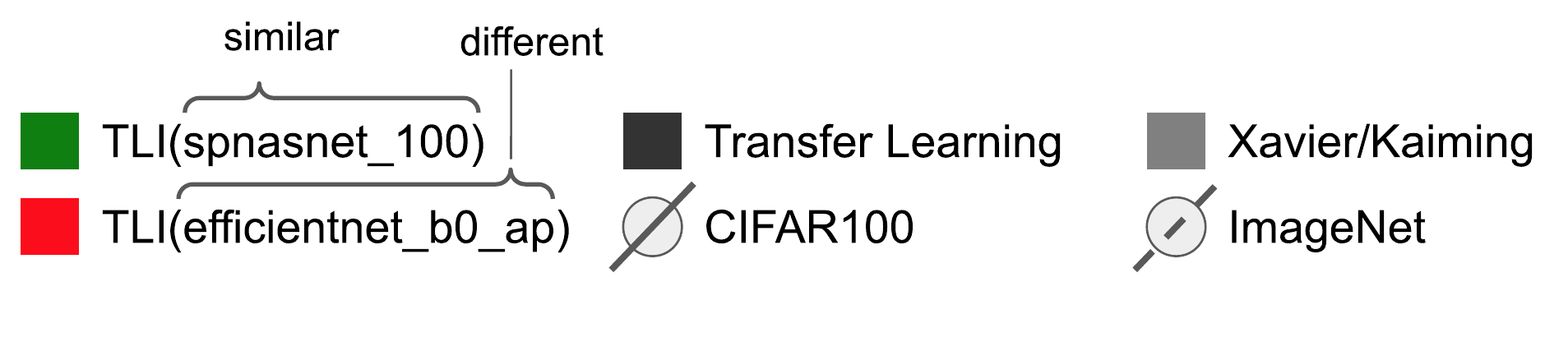}
	\vspace{-7mm}
	\center\includegraphics[width=\linewidth]{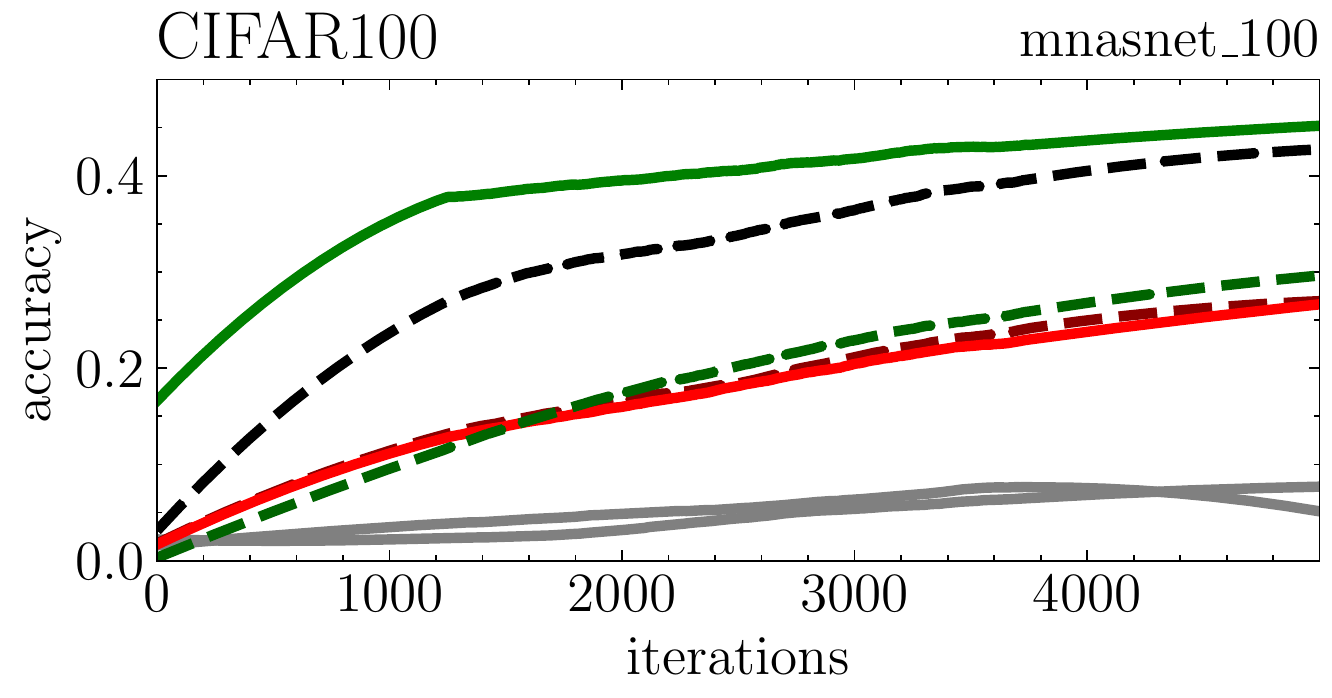}
    \vskip -5pt
	\caption{TLI (our) vs. standard methods. Training \textit{mnasnet\_100} on
	CIFAR100. Dashed line means that teacher was pre-trained on ImageNet.}
    \label{fig:tli_figure}
    \vspace{2mm}
    \center\includegraphics[width=\linewidth]{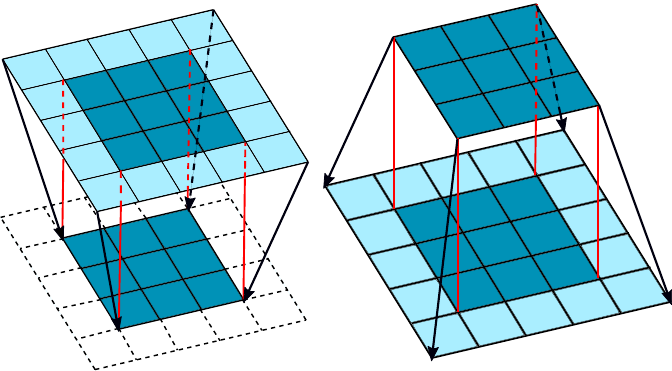}
    \vskip -5pt
	\caption{\ComboInjection: is a mix of ``center crop'' ($a$) and
	``resize'' ($b$), as tensor $x=\lambda a + (1 - \lambda)b$,
	where $\lambda$ is strength of interpolation.}
    \label{fig:da_figure}
	\vspace{-15mm}
\end{wrapfigure}


We propose a naive method of transferring knowledge between teacher and student
neural network: computationally cheap injection technique that does not require
any data samples.
The primary objective is to speed up the learning from scratch of a neural
network, if there is no previous pre-trained model.
We name this the TLI\footnote{Code: \url{https://github.com/maciejczyzewski/tli-pytorch}}
 (Transfer Learning by Injection) family of operations.
The work presented in this paper provides a minimal proof of concept. Further
research is required.

Student networks after transferring knowledge from teacher networks - that may
be different or pre-trained on different domains - are more likely to reach
convergence faster than the same student networks initialized with
Xavier/Kaiming methods.
Furthermore, a relationship exists between teacher-student similarity and
convergence times.
During the research, minor revisions to the architecture are made on a
continuous basis.
Typically, each model is designed to improve upon the previous model in some
way.
With our method, the models practically retain their previous performance and
continue to converge further.
There are a number of research workflows, including Kaggle Competitions, that
can be accelerated by the TLI method.

\newpage

The major contributions of this work:

\begin{enumerate}
	\item Presenting the algorithm for transferring parameters between
\underline{different} \underline{architectures} via computationally cheap
injection technique (does not require data) - drop-in replacement for
		Xavier/Kaiming initialization\citep{he2015delving}.
	\item Comparison of similarity between the current SOTA architectures
		(ImageNet), by utilising similarity score from presented method
		(Figure \ref{fig:matrix}).
\end{enumerate}


\section{Related work}

The term ``parameter remapping'' is used in \citep{fang2020fna++}, their work
describes an efficient framework for neural architecture search (FNA++).
Their method of transferring weights between different architectures is simple:
their weights are transferred by matching layers on depth, width and kernel
levels (crop center), which work only between same blocks.
Therefore, this method is insufficient for more complex architectures.

There is also Net2Net described in \citep{chen2015net2net} - in their work they
presents a simple method to accelerate the training of larger neural networks by
initializing them with parameters from a trained, smaller network.
The random mapping algorithm for different layers was done manually.
Developing a remapping algorithm would enable the Net2Net technique to be more
general.
This work further advances knowledge transfer by presenting a better remapping
technique that generalises prior methods.


\section{Weights Injection}

\begin{wrapfigure}{R}{.5\columnwidth}
	\vspace{-7mm}
	\center\includegraphics[width=\linewidth]{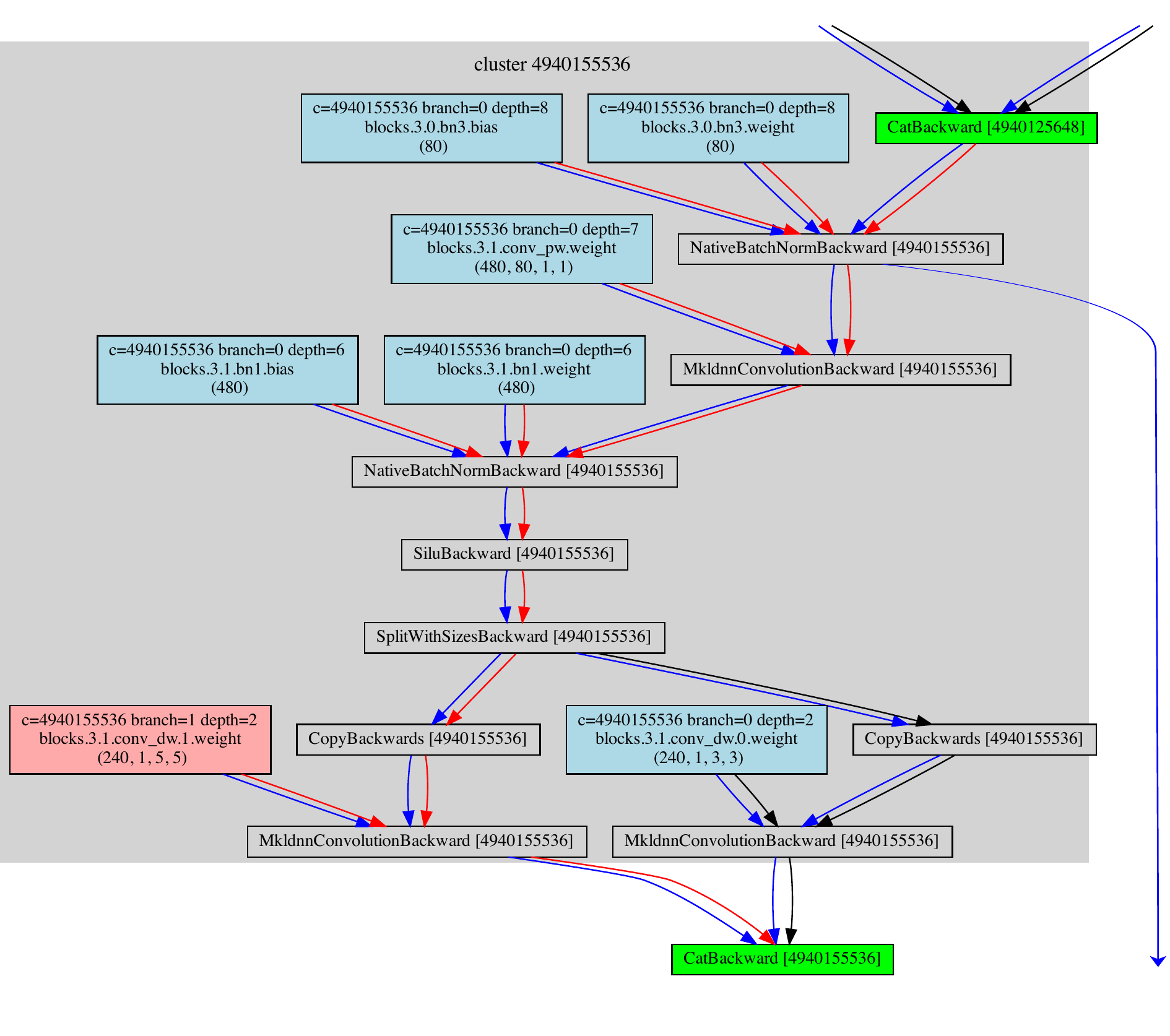}
	\vspace{-7mm}
	\vskip -5pt
	\caption{The tensor with a red background indicates the analyzed weight, a red path
	indicates the execution path, and a green block indicates the operation
	(\textit{CatBackward}).}
    \label{fig:path_algo}
	\vspace{-10mm}
\end{wrapfigure}

This algorithm uses two models as inputs: a teacher model to transfer knowledge
from, and a student model to transfer knowledge to.
Different architectures are recommended, unless you are dealing with classic
transfer learning (FT).
The presented method operates on an execution graph and can be applied to a
variety of tensor shapes.

The algorithm consists of two phases:
1) matching: the layers of the pre-trained model with the targeted model;
2) injection: the tensor is transformed into a desired shape.
No data samples are used in the algorithm, and there is no mutual loss between layers.
Moreover, the method can be extended to having multiple teachers, or to search for the best teacher from a library of pre-trained ones.
Several different architectures may make use of the same blocks as student.


\subsection{Matching method: path algorithm \& hashing}

Both models provided as input (student/teacher) are parsed in the following way:
1) the execution graph is clustered into submodules, divided by operations
\textit{AddBackward0}, \textit{MulBackward0},
\textit{CatBackward} defined in PyTorch \citep{NEURIPS2019}\footnote{PyTorch 1.7.0 notation};
2) for each tensor of weights, we need to find the path (list of operands) between one operation and another.
3) we iterate through the list of tensors of the model and the teacher, finding
the most similar execution path (using scoring function).
The Figure \ref{fig:path_algo} illustrates this process.

In practice, this algorithm has O(nm) complexity - where $n$ denotes the number of tensors containing the student weights, and $m$ is the same as $n$ but in the teacher model.
In this work, we will not discuss any ways of increasing speed.

The following is considered during the scoring comparison of the two execution paths:
depth; branch; used activations; submodule position from head; shape of
tensors.\footnote{This work is a draft, a thorough analysis of the formula and math will be presented in the final version.}

\subsection{Injection method: CenterCrop + Resize = \ComboInjection}

It is a combination of two operations: 1) resize to a new tensor size; and 2)
crop center, it does not modify the weights (teacher shape unchanged).
The strength of interpolation is controlled by the variable $\lambda$. Based on
empirical data, it is best when $\lambda$ is 0.75.
The operation has a desirable quality since it does not alter the weights for
the transferred tensor when the target and the input tensor have the same shape,
mimicking classic transfer learning (loading parameters).


\subsection{Multiple matches} 

In cases where we do not have a sure match, but a number of uncertain ones, the top K matches can be combined according to their weight according to the following:
\begin{equation}
\sigma (\mathbf {z} )_{i}={\frac {e^{z_{i}}}{\sum _{j=1}^{K}e^{z_{j}}}}{\text{ for }}i=1,\dotsc ,K{\text{ and }}\mathbf {z} =(z_{1},\dotsc ,z_{K})\in \mathbb {R} ^{K}
\end{equation}
\begin{equation}
	\mW_s^{(j)} = \sum _{i=1}^{K} \sigma (\mathbf {z} )_{i} T_i
\end{equation}

To calculate our transformation mixing function (2), we will use our $\mathbf{z}$ score vector in conjunction with softmax (1).


\section{Experiments}

\subsection{Datasets and implementation details}

Optimizer: Adam (lr=0.003); batch\_size=64; gradient accumulation (8 iterations) was used.
A single iteration is defined as one batch fit.
Every result is the average of three different runs from different seeds (series
are normalized with savgol\_filter).
Besides image normalization (std/var), we did not use augmentation.
Mixed precision is used for performance purposes.
These models and their weights have been imported from \citep{rw2019timm} library
``PyTorch Image Models''.

\subsection{Results and analysis}

The TLI requires further study and rigorous experiments.
The present work only involves simple experiments that can verify the proposed method only under some basic conditions.
More studies are needed.

\subsubsection{Initialisation on CIFAR100}

\begin{figure}[h]
\begin{center}
\center\includegraphics[width=\linewidth]{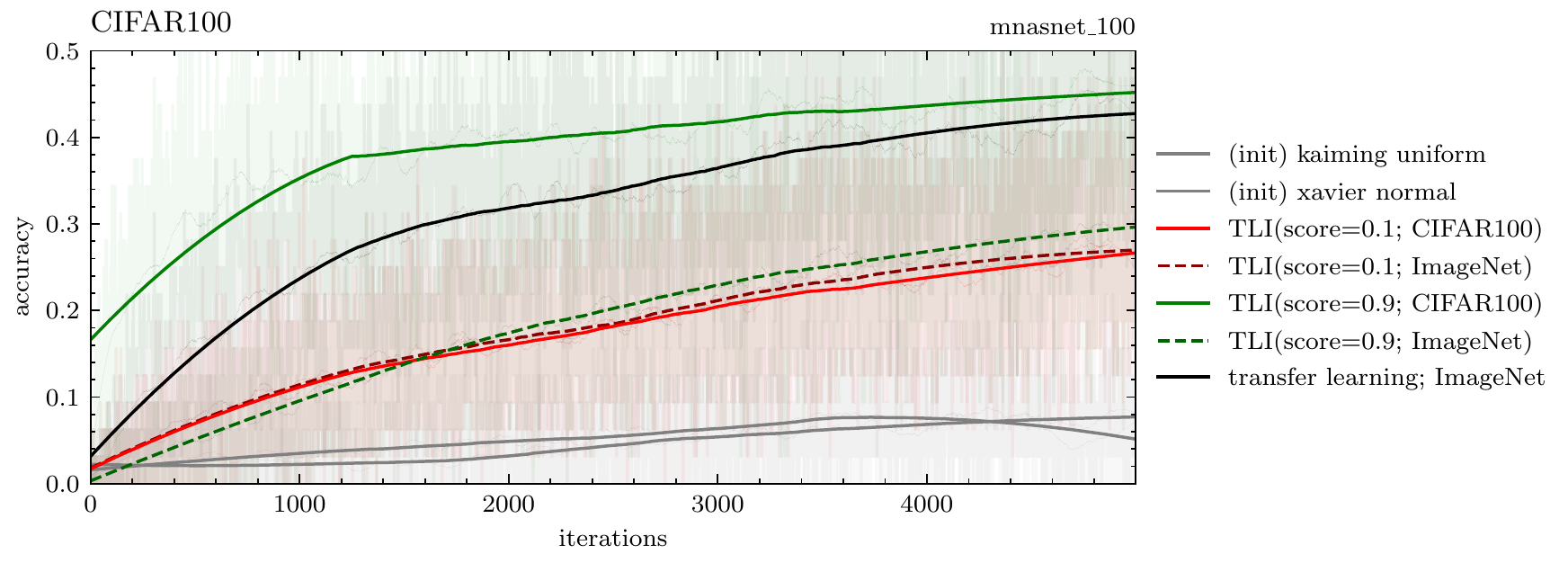}
\end{center}
\caption{
Compared with traditional initialization methods, TLI-based initialization converges faster after a few initial epochs (epoch=1.5k iterations).
When an architecture is new or layers have been modified, transfer learning (FT) is not applicable.
%
%
TLI(score=0.9) = \textit{spnasnet\_100};
TLI(score=0.1) = \textit{tf\_efficientnet\_b0\_ap}
(TLI scores are in Figure \ref{fig:matrix})
}
\label{fig:cifar100}
\end{figure}

Three different architectures were selected: mnasnet\_100, spnasnet\_100,
tf\_efficientnet\_b0\_ap.
We choose mnasnet\_100 as the base model to train on
CIFAR100 (results in Figure \ref{fig:cifar100}).
Knowledge was transfered from spnasnet\_100 (green) and
tf\_efficientnet\_b0\_ap (red) - each in two options:
a) pre-trained 5k iterations on CIFAR100 (normal line);
b) original weights from pre-trained models on ImageNet (dashed line).

Assuming that classical transfer learning is not applicable in this experiment
(black), we treat mnasnet\_100 as a new architecture that has never been
pre-trained before (for research purposes, we fine-tuned to compare).
As can be seen any TLI is better than Xavier/Kamming initialization.
A higher TLI score indicates that architectures are more similar to each other,
resulting in faster convergence.

\subsubsection{Without/with BatchNorm injection}

\begin{figure}[h]
	\begin{center}
	\subfloat[without]{
		\includegraphics[width=0.5\linewidth]{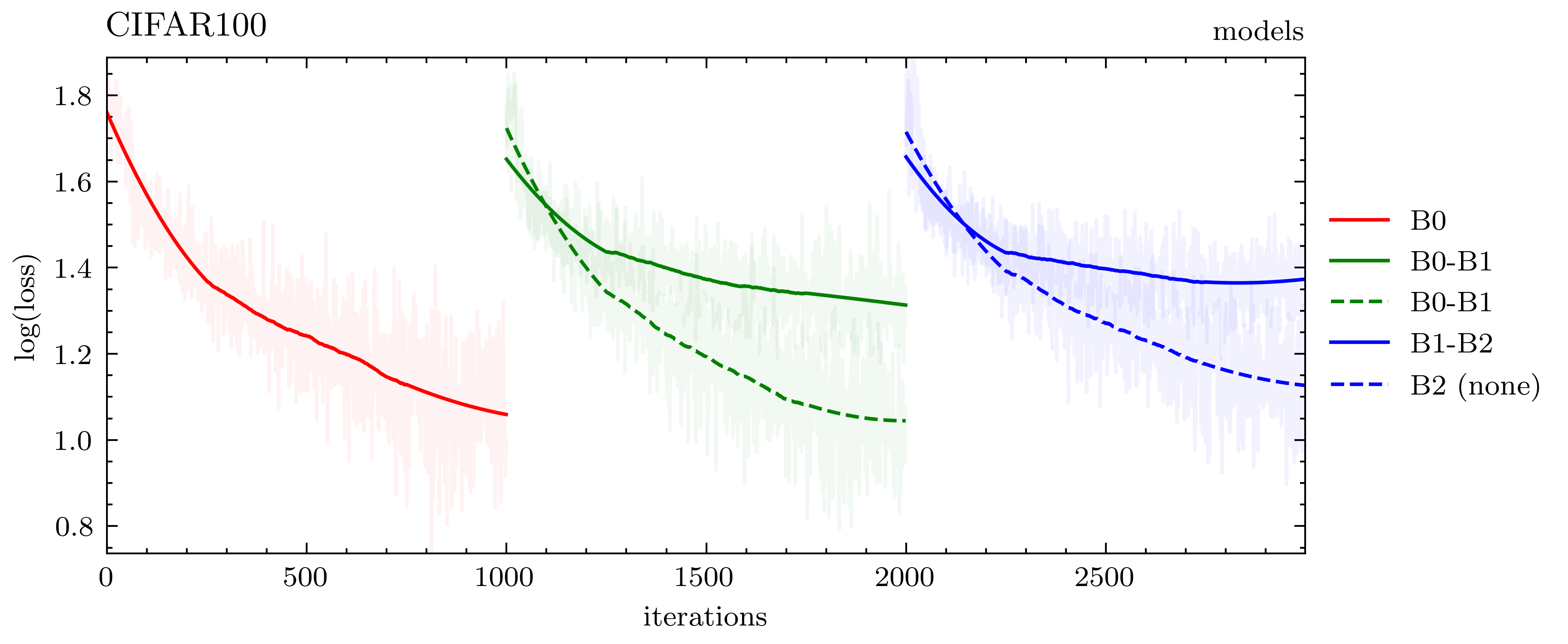}
	}
	\subfloat[with]{
		\includegraphics[width=0.5\linewidth]{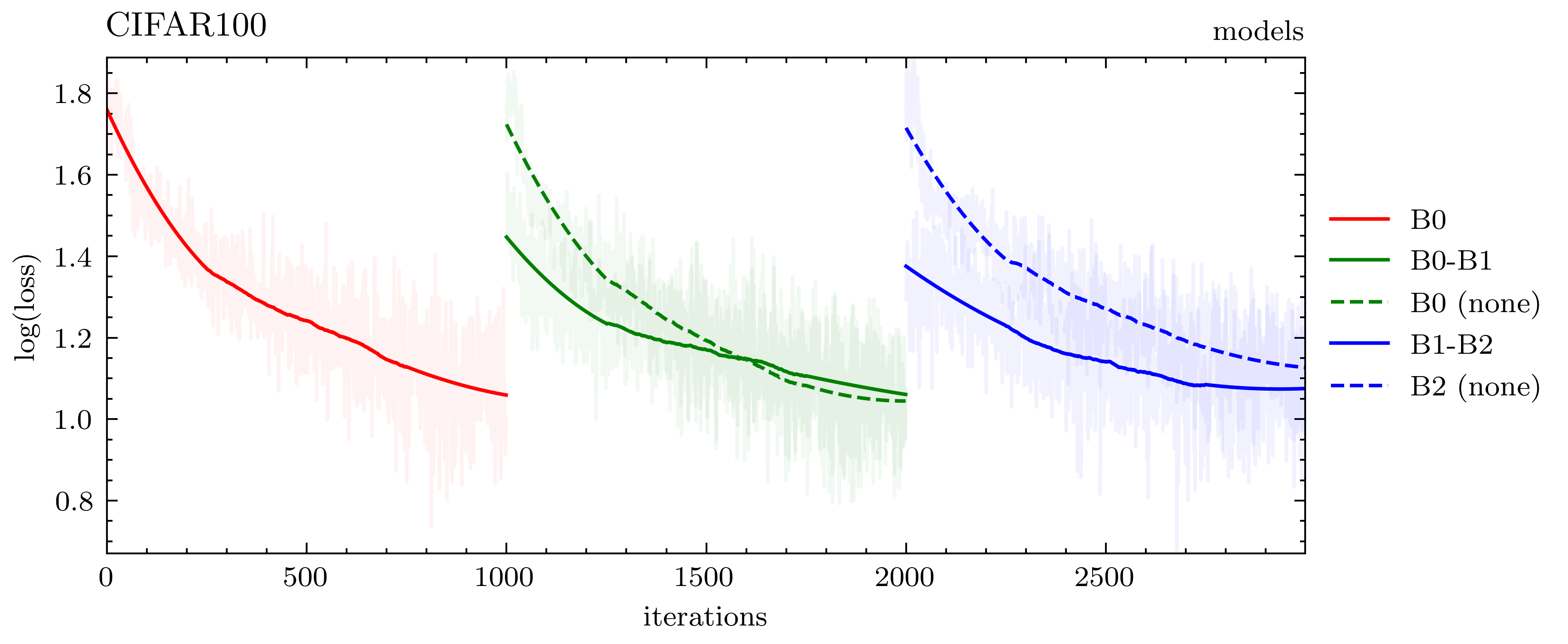}
	}
	\end{center}
	\caption{Comparison of impact of transferring BatchNorm weights.}
	\label{fig:batchnorm}
\end{figure}

This experiment will test whether it is worthwhile to transfer BatchNorm between different architectures (EfficientNet-B0, EfficientNet-B1, and EfficientNet-B2).
The EfficientNets described in \citep{tan2019efficientnet} were selected because they are similar in structure and block architecture (high TLI score).
Training each model involves 1000 iterations, following which TLI is applied to progressively larger models.
We will perform the first experiment with BatchNorm, followed by a second experiment without BatchNorm.

When architectures are similar or pre-trained using a task dataset, transferring
BatchNorm weights results in higher efficiency in most cases (in Figure
\ref{fig:batchnorm}).

\subsubsection{Use case: Kaggle Competitions} 

Many competitions use pre-trained models such as EfficientNet as a baseline
model. These models are then adapted for a new task (only output layers), and
trained on the competition dataset. This phase can be called fine-tuning (FT).
As a result, such a model, which has previously been trained on ImageNet, will
often adapt to competition very quickly (e.g. 100 epochs).
However, it is problematic to create a new architecture specifically for a
competition problem.
Because it requires a lot of research and it takes a lot of computational resources (training from scratch).
In certain situations, increasing the size of filters or strides improves the performance of tasks with high-resolution images.
Typically, manual weight assignment will make it unnecessary to undergo excessive training. This research proposes the TLI algorithm as an automated method of solving this problem.

\begin{figure}[h]
	\begin{center}
	\center\includegraphics[width=\linewidth]{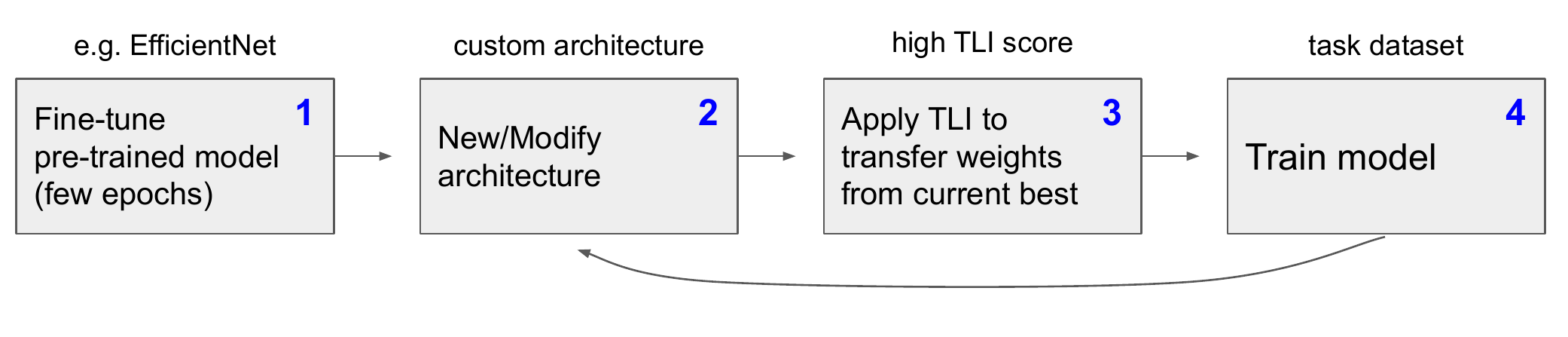}
	\end{center}
	\caption{Pipeline.}
	\label{fig:pipeline}
\end{figure}

This is the proposed pipeline:
\begin{enumerate}
\item train a few epochs model such as EfficientNet (generally, model with the
most similar TLI score to targeted model) on our task dataset.
This step can be omitted if there are public weights on Kaggle.
\item create a new or modify architecture with new features that will improve performance.
\item use TLI for transfer learning (from pre-trained model to new architecture).
\item train/fine-tune model, after a few epochs it accuracy should be equal to the
result of the pre-trained fine-tuned model (like EfficientNet) chosen as teacher
in step 1.
\end{enumerate}
When modifying activation function or one layer, the model should not lose
its accuracy from the very first epoch (repeat step 2/3/4 but as a teacher use
last trained model).


\section{Conclusions}

The hypothesis was tested whether it is better to transfer knowledge from any
architecture than to utilize Xavier/Kaiming as an initialization method.
It turned out that the presented technique converges faster, making it a drop-in
replacement.



\subsubsection*{Acknowledgments}
This work will be developed further in collaboration with Kamil Piechowiak and Daniel Nowak as part of a bachelor's thesis at the Poznan University of Technology, Poznan, Poland.

\bibliography{iclr2021_conference}
\bibliographystyle{iclr2021_conference}


\newpage
\appendix
\section{Appendix: similarity between architectures}

The table below presents similarity in range [0, 1], where 1 means that they are the
identical, while the score below 0.5 means that they are significantly different.
Clearly, models like
\textit{tf\_efficientnet\_lite0} (tensorflow weights) and
\textit{efficientnet\_lite0} give same results. Architectures like RegNet
family is substantiality different then ResNet alternatives.

\begin{figure}[h]
	\begin{center}
	\center\includegraphics[width=\linewidth]{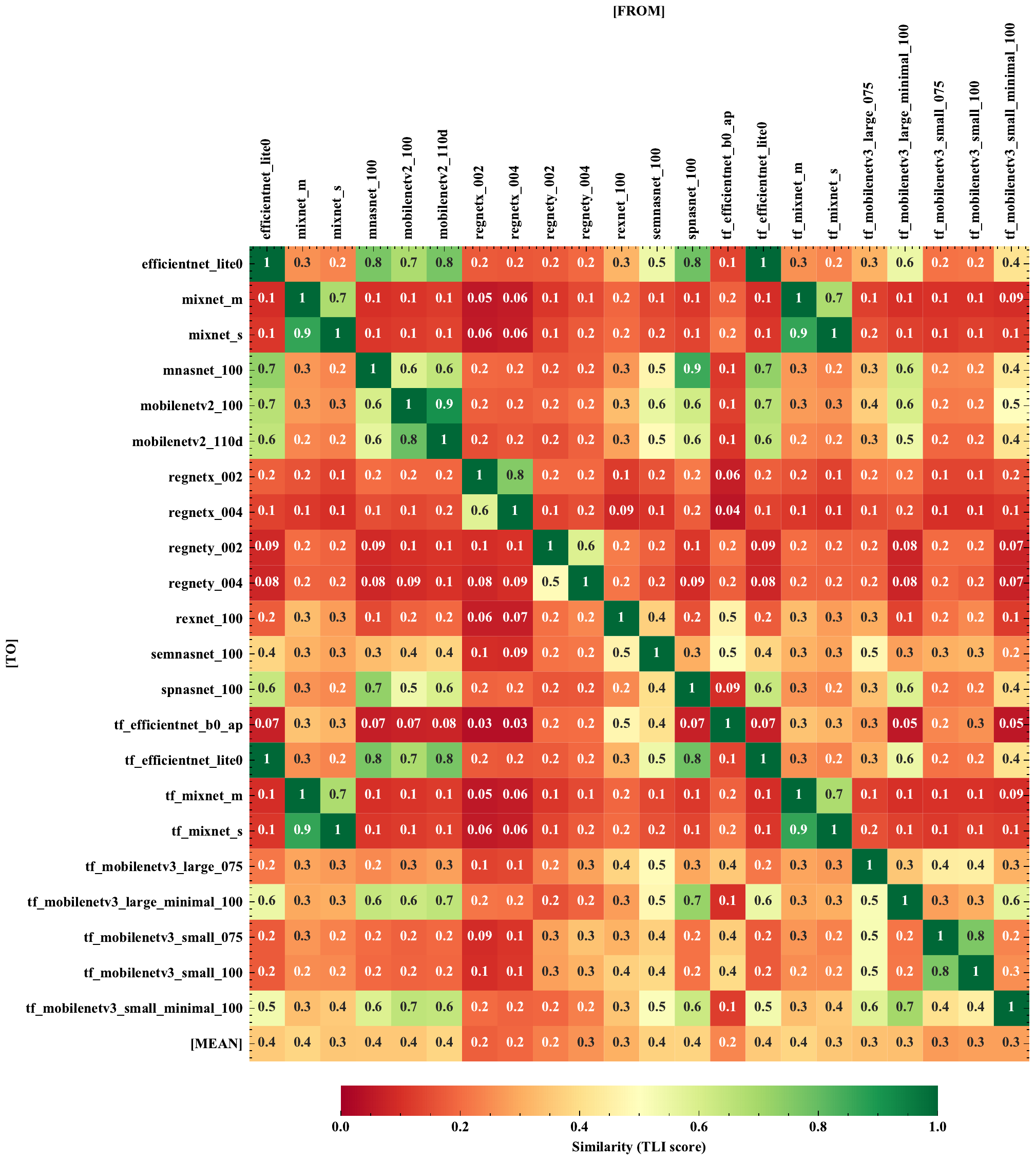}
	\end{center}
	\caption{Similarity between architectures.}
	\label{fig:matrix}
\end{figure}

\end{document}

%% file: math_commands.tex
\usepackage{amsmath,amsfonts,bm}

\def\eqref#1{equation~\ref{#1}}

\def\1{\bm{1}}

\def\mW{{\bm{W}}}

\DeclareMathAlphabet{\mathsfit}{\encodingdefault}{\sfdefault}{m}{sl}
\SetMathAlphabet{\mathsfit}{bold}{\encodingdefault}{\sfdefault}{bx}{n}

